\documentclass[10pt,twocolumn,letterpaper]{article}

\usepackage{cvpr}              % To produce the CAMERA-READY version
\usepackage[british]{babel}
\usepackage[utf8]{inputenc}
\usepackage{babelbib}
\usepackage{url}
\usepackage{graphicx}
\usepackage{subcaption}
\usepackage{calc}
\usepackage{floatflt}
\usepackage{amssymb, amsmath, amsthm}
\usepackage{tabularx}
\usepackage{multirow}
\usepackage{array}
\usepackage{xcolor}
\usepackage{soul}
\usepackage{booktabs}
\usepackage{bm}
\usepackage{stmaryrd}
\usepackage{stackrel}
\usepackage{algpseudocode}
\usepackage{algorithm}
\usepackage{rotating}
\usepackage{mwe,tikz}
\usepackage[percent]{overpic}
\usepackage{float}
\usepackage{listings}

\usepackage{numprint}
\usepackage{memhfixc}
\usepackage{makecell}
\usepackage{textcomp}
\usepackage{lineno}
\definecolor{cvprblue}{rgb}{0.21,0.49,0.74}
\usepackage[pagebackref,breaklinks,colorlinks,allcolors=cvprblue]{hyperref}

\def\paperID{4955} % *** Enter the Paper ID here
\def\confName{CVPR}
\def\confYear{2025}

\title{SimCMF: A \underline{Sim}ple \underline{C}ross-\underline{m}odal \underline{F}ine-tuning Strategy\\from Vision Foundation Models to Any Imaging Modality}

\author{Chenyang Lei$^{1,2}$\thanks{Equal contribution. $\dagger$ Corresponding authors.} \quad Liyi Chen$^{1,3}$\footnotemark[1] \quad Jun Cen$^4$ \quad Xiao Chen$^{1,3}$ \\ Zhen Lei$^{1,5}$ \quad Felix Heide$^2$ \quad Qifeng Chen$^{4\dagger}$ \quad Zhaoxiang Zhang$^{1,5\dagger}$ \\
$^1$Center for Artificial Intelligence and Robotics, HKISI, CAS 
$^2$Princeton University\\
$^3$The Hong Kong Polytechnic University
$^4$The Hong Kong University of Science and Technology \\
$^5$State Key Laboratory of Multimodal Artificial Intelligence Systems, CASIA
}

\begin{document}
\maketitle
\newcommand{\exptrainingsctrach}{22.15}
\newcommand{\expSimCMF}{53.88}

\newcommand{\mattrainingsctrach}{25.43}
\newcommand{\matrandomly}{58.89}
\newcommand{\matlinear}{63.96}
\newcommand{\mattranspose}{70.90}
\newcommand{\matours}{72.69}

\begin{abstract}
Foundation models like ChatGPT and Sora that are trained on a huge scale of data have made a revolutionary social impact. However, it is extremely challenging for sensors in many different fields to collect similar scales of natural images to train strong foundation models. To this end, this work presents a simple and effective framework, SimCMF, to study an important problem: cross-modal fine-tuning from vision foundation models trained on natural RGB images to other imaging modalities of different physical properties (e.g., polarization). In SimCMF, we conduct a thorough analysis of different basic components from the most naive design and ultimately propose a novel cross-modal alignment module to address the modality misalignment problem. We apply SimCMF to a representative vision foundation model Segment Anything Model (SAM) to support any evaluated new imaging modality. Given the absence of relevant benchmarks, we construct a benchmark for performance evaluation. Our experiments confirm the intriguing potential of transferring vision foundation models in enhancing other sensors' performance: SimCMF can improve the segmentation performance (mIoU) from 22.15\% to 53.88\% on average for evaluated modalities and consistently outperforms other baselines. The code is available at \url{https://github.com/mt-cly/SimCMF}.
\end{abstract}

\section{Introduction}

\begin{figure}[t]
\raggedright
\includegraphics[width=\linewidth]{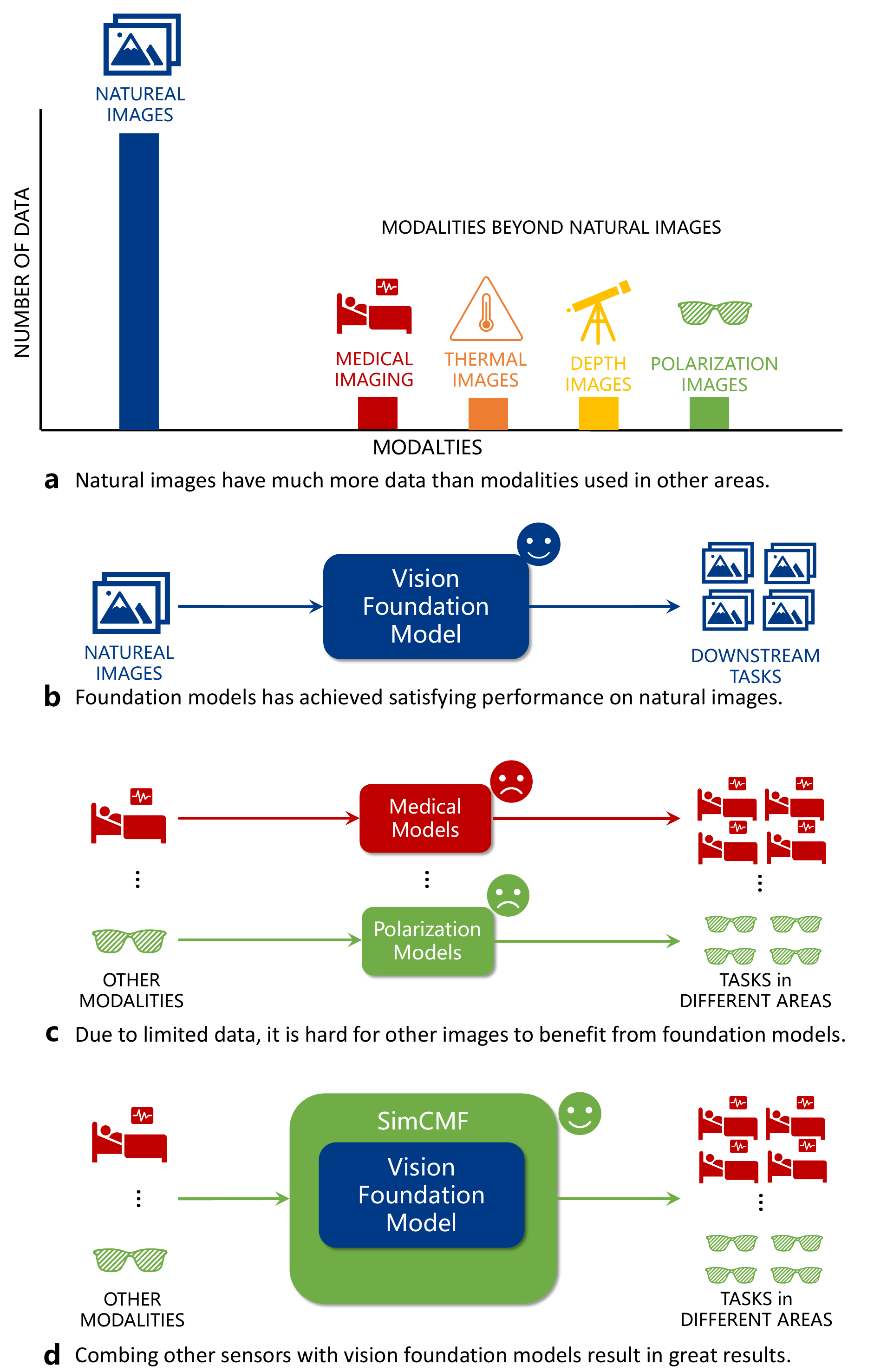}

\caption{\textbf{Transferability Across Modalities.} \textbf{a}, the number of natural images is significantly larger than images in other modalities in different areas, including medical imaging, thermal images, depth images, and polarization images. \textbf{b}, natural images can train vision foundation models, which can be applied to achieve strong performance on different downstream tasks. \textbf{c}, it is very challenging for other modalities to benefit from training foundation models due to limited data. \textbf{d}, our proposed SimCMF explores the transferability from the pretrained vision foundation model to different imaging modalities. }
\label{fig:framework}
\end{figure}

% What is the problem? Why is it important?
Foundation models have revolutionized computer vision~\cite{kirillov2023segment, bai2024sequential,bommasani2021opportunities} and natural language processing~\cite{kenton2019bert,brown2020language} across the fields, from personal assistance to self-driving vehicles and medical diagnosis~\cite{ma2024segment,zhou2024pre,yang2024vision,gehrig2024low,yako2023video,merchant2023scaling}. Diverse downstream tasks rely directly or indirectly on foundation models by finetuning foundation models that are pretrained on large-scale data with pretext tasks~\cite{pan2009survey}. However, while diverse types of sensors~\cite{wu2023medical,huang2023polarization,lei2022shape,lei2020polarized,sun2022seeing,gallego2020event,dong2016hyperspectral,tseng2021neural} are applied in various domains in the world, e.g., medical imaging, robotics, and fundamental science, not all of them benefit from the development of foundation models. 
This is because it is challenging for other sensors~\cite{wang2024sub,mao2024multimodal} to collect large-scale training data like natural images, as shown in Figure~\ref{fig:framework}. 

This work explores the following problem: transferring the vision foundation models to modalities other than natural images. While training foundation models and fine-tuning them on downstream tasks has been extensively studied~\cite{ye2024superanimal,pai2024foundation,cai2024pretrainable}, the potential of generalizing foundation models to novel imaging modalities is not fully explored. Arguably, transferring the foundation models to various input modalities like task transfer learning 
has the potential to unleash the power of the foundation model on specific sensors: we can utilize the advantages of sensors 
in capturing specific physical properties of objects in the world with a strong foundation model. 

% Why is it challenging?
The challenges for transferring vision foundation models to other imaging modalities come from two sides: the modality misalignment and the fine-tuning cost. A key challenge of cross-modality fine-tuning comes from the modality gap: the captured physical signals and the data representation can be highly different, such as the dimensions, the dynamic ranges, and semantic information. Among many differences, the dimension misalignment is one of the major challenges, preventing people from fine-tuning on new modalities directly. A simple example is that RGB images capture the visible color of objects with three channels. In contrast, a polarization sensor can capture the polarization state of light with more than three channels. The second challenge comes from the fine-tuning cost, which is increasing rapidly along with the quick growth of the model size of foundation models. To this end, a systematical analysis for applying different parameter-efficient fine-tuning strategies to cross-modal fine-tuning can be beneficial.

% Existing work.
Researchers have attempted to explore cross-modal fine-tuning in different modalities, but most works focus on transferring a pretrained modality to another specific modality, including from language to vision~\cite{lu2022frozen,dinh2022lift} or protein sequences~\cite{vinod2023reprogramming}, from natural images to medical imaging~\cite{wu2023medical,ma2023medsam}. A line of literature has studied how to design a general cross-modal fine-tuning framework for different modalities~\cite{lu2022frozen,shen2023cross}. However, they do not carefully handle the modality misalignment~\cite{lu2022frozen}, require large computational cost~\cite{wu2023medical}, or requiring extra data~\cite{shen2023cross}. Besides, they do not take into account fine-tuning strategies, which is quite important in practice. In contrast, we study how to transfer the vision foundation model comprehensively, including handling modality misalignment and analyzing fine-tuning strategies.

% Our contribution
% SimCMF is designed to accept any imaging modality as input for fine-tuning, which does not require domain-specific knowledge, such as the relationship between the modality and natural images. 
To investigate this problem, we introduce SimCMF: a simple framework for cross-modal fine-tuning from a vision foundation model to any imaging modality. SimCMF consists of a cross-modal alignment module and a pretrained foundation model backbone. We first propose a simple and effective cross-modal alignment module to solve the modality misalignment problem between the target modality and the pretrained vision modality. We start our exploration from the most naive design, a randomly initialized embedding layer, and then improve it gradually with several basic components. In our exploration, we identify the key components for cross-modal alignment. Secondly, we further provide a comprehensive empirical study of fine-tuning strategies on cross-modal fine-tuning, including full-fine-tuning (FFT) and parameter-efficient fine-tuning (PEFT) strategies like LoRA~\cite{hu2021lora}, MLP Adapter~\cite{houlsby2019parameter}, prompt tuning~\cite{zhu2023visual}. Results confirm the potential of PEFT for cross-modal, which is consistent with the observations in unimodal fine-tuning.

In this paper, we apply SimCMF to a representative vision foundation model Segment Anything Model (SAM)~\cite{kirillov2023segment} so that it can be used for segmentation in different image modalities. SAM is trained on 11 million images for a fundamental image segmentation task. To enable a fair comparison to study the transferring performance of cross-modal fine-tuning SAM on novel modality, we build a dedicatedly designed segmentation benchmark that consists of datasets captured by various types of sensors, including polarization sensors, depth sensors, thermal sensors, and other types of sensors.

Extensive results demonstrate that SimCMF can achieve significant performance improvement across image modalities compared with models that are trained on specific modalities only. We find that SimCMF does improve the performance of other modalities despite these sensors capturing different physical properties in different representations. We hope our SimCMF can serve as a flexible and solid tool for transferring vision foundation models to other image modalities in different areas.

\begin{figure*}
\vspace{-15pt}
\centering
 \includegraphics[width=0.7\linewidth]{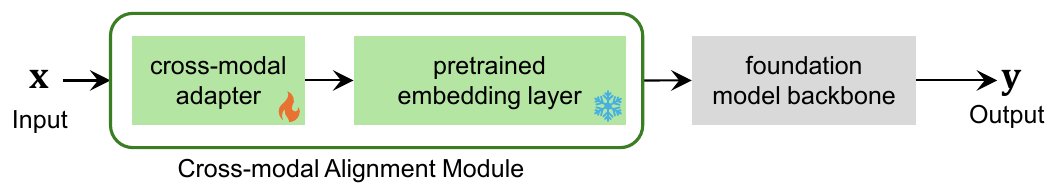}    
\caption{\textbf{SimCMF Conceptual Overview.} SimCMF receives new modality  $\mathbf{x}$ as input and pass it through a cross-modal alignment module to obtain an embedding. The embedding matches the dimension of a pretrained foundation model backbone, and then we obtain the output $\mathbf{y}$. The input and foundation are designed in a generic formulation for different input modalities and foundation models. In this work, we select SAM as a representative foundation model for a detailed study. }
\label{fig:SAM-AIM}
\end{figure*}

\section{Related Work}

\noindent \textbf{Unimodal fine-tuning.} Unimodal transfer learning is commonly used in computer vision, which first pretrains the model to learn prior knowledge and then fine-tunes it on another downstream task. It has shown to be effective in various areas~\cite{zamir2018taskonomy}. For example, the models are first trained on large-scale datasets~\cite{deng2009imagenet} in a contrastive learning~\cite{chen2020simple, moco, moco-v2, moco-v3, BYOL} or a masking inpainting way~\cite{he2022masked, simmim, beit} and then used for the downstream task with RGB image input. Besides, the pertaining model works well in other scenarios like predictions of RNA secondary structure~\cite{singh2019rna}, metal-organic framework~\cite{kang2023multi, pan2023transfer} and fault slip~\cite{wang2021predicting}.

\vspace{\baselineskip}

\noindent\textbf{Cross-modal fine-tuning.} 
The modalities suffering from limited training data fail to perform the pretrain-tuning paradigm. Cross-modal fine-tuning is a potential way to solve this problem. However, most research explores this problem in a modality-specific style for a specific pair of modalities. For example, Radhakrishnan et al.\cite{radhakrishnan2023transfer} study transfer learning on image classification and virtual drug screening applications. Many works~\cite{lu2022frozen,dinh2022lift,llava,li2024llava} study the transferability from language models to vision. Zhang et al.~\cite{zhang2024generalist} employ the vision-language foundation model for biomedical tasks. Vinod et al. ~\cite{vinod2023reprogramming} attempts to apply language models to protein sequences~\cite{vinod2023reprogramming}. Wu et al.~\cite{wu2023medical} and Ma et al.~\cite{ma2023medsam} attempt to transfer the vision segmentation model to medical imaging. 
% For example, modalities like polarization, structured light~\cite{choi2023neural}, and event camera~\cite{gehrig2024low,guo2024eventlfm}, which have proven instrumental in 3D imaging and auto-driving, but there is no large scale data for pertaining. 

Few works~\cite{lu2022frozen} study a modality-agnostic cross-modal workflow of transferring the knowledge from pretrained data to downstream tasks. Lu et al.~\cite{lu2022frozen} proposes a framework FPT for transferring pretrained transformers to different inputs. However, they only study pretrained language models in a frozen way. Most recently, Shen et al.~\cite{shen2023cross} propose a general cross-modal transfer learning framework for diverse modalities, including language, vision, tabular, $etc$. Nevertheless, they do not take into account finetuning strategies, which is quite important in practice. In contrast, we explore the transferability of the vision foundation model by investigating modality misalignment and finetuning strategies comprehensively.

\vspace{\baselineskip}

\noindent\textbf{Domain adaption.} Domain adaptation~\cite{da_adverse_learning_1,da_adverse_learning_2,da_adverse_learning_3,da_self_learning_1,da_self_learning_2,da_self_learning_3} aims to transfer source domain knowledge to the target domain. Heterogeneous domain adaptation~\cite{hda_1,hda_2} extensively discusses the feature space change. However, they usually assume the source training data is available.

% Different from these methods considering the domain gap in RGB images, we try to alleviate the modality gap between RGB images and other image modalities, it is more challenging since the physical forward model changes between the image channels.
% While 

\vspace{\baselineskip}

\noindent \textbf{Parameter-efficient fine-tuning.} Parameter-efficient fine-tuning is also closely related to our research. Fully fine-tuning a large transformer model costs large GPU memory and training time. Parameter-efficient fine-tuning solves this problem by freezing the pretrained foundation model and only fine-tuning a small number of parameters, which has been shown to achieve comparable or even better performance than full fine-tuning. It was first proposed in the natural language processing task~\cite{houlsby2019parameter,hu2021lora,li2021prefix} and then explored in the computer vision task~\cite{chen2022adaptformer,jia2022visual}. 

% Visual prompt tuning~\cite{jia2022visual} adds some learnable tokens before each transformer block. 

% Visual adapter~\cite{chen2022adaptformer} inserts small multilayer perceptrons (MLPs) to the feed-forward network in a residual way~\cite{he2016deep}. 

% Prefix-tuning~\cite{li2021prefix} adds a few parameters before each multi-head attention layer. 
% LoRA~\cite{hu2021lora} optimize the rank-decomposition matrices of layers' change and achieve zero inference latency. 

\vspace{\baselineskip}

\noindent \textbf{Modality-agnostic architectures.} Unified architectures and learning algorithms for various data modalities have emerged recently. Designing a foundation model~\cite{bommasani2021opportunities, internimage, llava, imagebind} for various modalities becomes a goal for the community. The transformer architecture~\cite{vaswani2017attention} has been proven to be very effective in different domains, including natural language~\cite{kenton2019bert,brown2020language,touvron2023llama}, vision~\cite{dosovitskiy2020image,liu2021swin,wang2021not}, point clouds~\cite{guo2021pct,zhao2021point,wu2022point}, audio~\cite{gong2021audio,chen2022hts,verma2021audio}, and so on. Recently, architectures~\cite{jaegle2021perceiver,jaegle2021perceiverio,zhang2023meta,team2024chameleon} are proposed for the general perception of various types of data modalities. Tamkin et al.~\cite{tamkin2021dabs} construct a benchmark for domain-agnostic self-supervised learning algorithms~\cite{wu2023randomized,wang2023internimage}, including natural images, language, and sensors. Our SimCMF is also designed to handle different imaging modalities in a modality-agnostic formulation.

% Meta-transformer~\cite{} uses a unified encoder for 12 modalities with each modality using a specific encoding way. 
% PerceiverIO~\cite{} unify the input and output structures, which demonstrate the effectiveness of images and videos. 
\section{SimCMF: Simple Cross-modal Fine-tuning }

%SimCMF
% We seek to minical revision.
% MAT:
% - Motivation
% - How do we design it
% - Details
% Foundation model
% - Designed for different foundation models.
% - Keep foundation models and output as the same for minical revision
% - SAM is representative 

In cross-modal fine-tuning with any imaging modality, the target data are captured from different imaging sensors, such as depth sensors, polarization sensors, thermal sensors, and so on. As it is quite challenging for these sensors to capture similar scales of data like natural RGB images, fine-tuning a vision foundation model with limited target modality data is desirable. However, it is non-trivial to design a unified fine-tuning architecture that satisfies different modalities as they have very different features, such as resolution, dimensions, and the captured physical signals.

In this section, we introduce SimCMF, a simple cross-modal fine-tuning framework to explore the transferability of vision foundation models to other imaging modalities and address the above challenges. As shown in Fig.~\ref{fig:framework}, our framework is inspired by the attractive performance of unimodal fine-tuning, from a pretrained task to different downstream tasks. SimCMF consists of a cross-modal alignment module and a foundation model, as illustrated in Figure~\ref{fig:SAM-AIM}. The cross-modal alignment module is designed to align the target modality with different dimensions with the original vision modality. For the foundation model backbone, we apply different fine-tuning strategies to study this problem, including parameter-efficient fine-tuning (PEFT) and full fine-tuning (FFT).  

% SimCMF is designed to accept any imaging modality as input for transfer learning. It does not require domain-specific knowledge, such as the relationship between the modality and natural images. 
% Given an input $\mathbf{x}$, the output $\mathbf{y}$ is obtained by:
% \begin{align}
%     \mathbf{y} = f(\mathbf{e}) = f(m(\mathbf{x})),
% \end{align}
% where $\mathbf{e}$ is the output embedding of our MAT layer. 

\subsection{Network architecture} 
\noindent\textbf{Cross-modal alignment module.}
 Different from unimodal fine-tuning, novel image modalities captured from alternative sensors may have different dimensions, which prevents us from using existing fine-tuning strategies directly. Considering a new input data that is captured from a novel sensor. The dimension of the new data $C$ is a modality-specific value (e.g., $C=1$ for a depth image, $C=9$ for a polarization image), which might be different from the RGB image ($C=3$).

\begin{itemize}
    \item \textbf{Pretrained embedding layer.} The pretrained embedding layer receives RGB images with three channels and outputs $d$ channel features. However, many imaging modalities cannot use the embedding layer directly due to dimension misalignment. To solve this problem, a naive approach is randomly initializing a new modality embedding layer and training the whole network jointly. In our experiments, we observe that using the \textit{frozen pretrained embedding layer} is key to obtaining satisfying fine-tuning performance, even if the signals captured from the new modality are quite different from the natural images. 
    
    \item \textbf{Cross-modal adapter.} To use the pretrained embedding layer, it is necessary to handle this dimension misalignment problem. Using a linear layer to change the dimension is commonly used in different tasks. However, we observe that they are suboptimal and we design a cross-modal adapter after conducting extensive experiments, which is simple but effective. Specifically, our cross-modal adapter consists of $l$ convolution layers with $k$ kernel size and activation functions after each convolution layer for nonlinearity except for the last layer. In our experiments, we observe that $l=2$ and $k=3$ results in the best performance. Note that when $l=1, k=1$ the cross-modal adapter equals to a simple $1\times1$ convolution layer (i.e., linear layer). Besides, when $l=2, k=1$, this projector is a commonly used MLP layer for changing the dimension in contrastive learning and vision language models~\cite{chen2020simple,moco-v3,li2024llava}. We observe that replacing linear layers with convolution layers and adding nonlinearity are beneficial to aligning two imaging modalities. Besides, we notice that using too many layers leads to unstable training, which might be because the initialization affects the performance a lot.
\end{itemize}

\vspace{\baselineskip}
\noindent\textbf{Foundation model backbone.} We seek to modify the foundation model minimally compared to the pretraining stage. Our cross-modal alignment module receives a C-dimensional to an embedding with $d$ dimensions, which is used as the input to the foundation model backbone. We load the pretrained weights from the vision foundation model directly. We will add a few trainable parameters if we use the parameter-efficient fine-tuning strategies, similar to unimodal fine-tuning. In this paper, we focus on fine-tuning the model for different modalities, and thus we keep the output head the same as the pretrained foundation model. 

% Specifically, in our experiments, we select SAM as the representative vision foundation model $f$. The original module in SAM receives a three-dimensional RGB image to an embedding with $d$ dimensions. 

\subsection{Training} 
\textbf{Training objective.} Given a dataset that consists of images captured from a new image sensor, we train the model using the same loss function of the vision foundation model as we focus on finetuning the model for different modalities. Hence, we keep the output head the same as the pretrained foundation model. 
% In this paper, we use the loss proposed in SAM to fine-tune the whole network architecture.
\vspace{\baselineskip}

\noindent\textbf{Finetuning strategies.} While there are analyses for specific domains, such as medical imaging~\cite{ma2023medsam}, to the best of our knowledge, there is no systematic analysis for cross-modal fine-tuning for different imaging modalities. Hence, in our experiments, we investigate different fine-tuning strategies, including LoRA~\cite{hu2021lora}, MLP Adapter~\cite{chen2022adaptformer}, full fine-tuning, $etc$. We analyze the relationship between the number of data, best learning rate, and peak performance for these finetuning strategies. Details are presented in Section~\ref{sec:finetuning}.

% Detailed results and analysis are presented in Section~\ref{sec:explore}. Note that we are not designing a \textit{multi-modal} foundation model that can process different modalities at the same time. Instead, we finetune the model on \textit{each modality}. 

% \newcommand{\exptrainingsctrach}{22.15}
% \newcommand{\expSimCMF}{53.88}

% \newcommand{\mattrainingsctrach}{25.43}
% \newcommand{\matrandomly}{58.89}
% \newcommand{\matlinear}{63.96}
% \newcommand{\mattranspose}{70.90}
% \newcommand{\matours}{72.69}

\section{Experiments}

\subsection{Experimental setup} 
\paragraph{Evaluated foundation model.} Visual foundation models have developed very fast~\cite{xdecoder,4m,seem,wang2023internimage,rombach2021highresolution}, and this paper selects Segment Anything Model (SAM)~\cite{kirillov2023segment} as a backbone for exploring experiments as it is one of the most representative foundation models in computer vision. 

SAM has three components: an image encoder, a prompt encoder, and a mask decoder. The image encoder receives image patches as input and computes image features. The prompt encoder embeds prompts, $i.e.$, points, boxes, text, or masks. Both image features and prompt embedding are fed into a lightweight mask decoder to obtain mask predictions.
The released SAM model is trained on the large-scale SA-1B dataset, which contains over 1 billion
automatically generated masks (400× more masks than any existing segmentation datasets)
and 11 million images. Several works~\cite{ji2023sam, tang2023camouflage, chen2023samfail, ma2023medsam} focus on adapting the SAM to different domains of RGB images, while we use SAM as the vision foundation model to explore the modality transfer task. Although some works~\cite{wu2023medical, cen2023sad} have discussed the SAM adaption with specific modality ($e.g.$, MRI, depth), we are toward a more general setting handling an arbitrary modality.

% In our experiments, we investigate different finetuning strategies, including LoRA~\cite{hu2021lora}, MLP Adapter~\cite{chen2022adaptformer}, full finetuning, $etc$. We train the model on each modality respectively. We implement the SimCMF using PyTorch. More details for the training can be found in the supplementary materials.

% Note that we are not designing a \textit{multi-modal} foundation model that can process different modalities at the same time. Instead, we finetune the model on \textit{each modality}. 

% as alternative baseline approach do not exist

% Combing the gate operation with learnable embedding branch improves the performance from 69.48 to 72.14 on the evaluate dataset. 

% With above exploratory experiments, we propose a compact {MAT} layer. Figure~\ref{fig:MAT} presents the details of MAT. This module $m$ requires a pretrained RGB embedding layer $m_v$ and the novel modality dimension $C$ as input. The dimension of output modality embedding is $d$. It consists of two branches: frozen vision embedding branch and learnable embedding branch. The frozen one is simply a $1\times1$ convolution layer with a frozen vision embedding layer. The learnable embedding branch is added as residual by a zero-initialization $1\times1$ convolution layer. Besides, we note that it is important to add a gate operation before two branches. Otherwise, these two branches tend to learn similar features close to vision embeddings. Combing the gate operation with learnable embedding branch improves the performance from 69.48 to 72.14 on the evaluate dataset. 

\begin{table}[t]
    \centering
    \resizebox{.7\linewidth}{!}{
        \begin{tabular}{lcc}
            \hline
            \multirow{2}{*}{Method} & {Training } & {Training with} \\ 
            & {from scratch} & {SimCMF} \\ \hline
            {Thermal} & 32.08 & \textbf{57.25} \\
            {Polarization} & 25.43 & \textbf{72.69} \\
            {Depth} & 22.38 & \textbf{45.09} \\
            {HHA} & 22.89 & \textbf{44.02} \\ 
            {Near Infrared} & 7.99 & \textbf{50.36} \\ \hline
            {Average} & 22.15 & \textbf{53.88} \\ \hline
        \end{tabular}
    }
    \caption{\textbf{Performance Evaluation on Different Modalities.} The proposed method SimCMF improves the segmentation performance significantly on all evaluated modalities compared with training the models from scratch. Specifically, SimCMF improves the mIoU from 22.15\% to 53.88\% for all evaluated modalities on average on our constructed AIMS benchmark.}
    \label{tab:comparison_a}
\end{table}

\begin{table}[t!]
    \centering
    \resizebox{1.0\linewidth}{!}{
    \begin{tabular}{lccccc}
    \hline
        Method &  RGB-T & RGB-D & RGB-HHA & RGB-NIR \\ \hline
        Train from scratch & 43.17 & 24.02 & 22.95 & 10.25 \\ 
        Zero-shot &  48.76 & 49.03 & 49.03 & 44.66 \\ %\hline
        SimCMF &   \textbf{85.29} & \textbf{57.73} & \textbf{57.25} & \textbf{55.81} \\ \hline
    \end{tabular}

    % \begin{tabular}{lccccc}
    % \hline
    %     Method &MAT &  RGB-Thermal & RGB-Depth & RGB-HHA & RGB-Near Infrared \\ \hline
    %     Train from scratch & - & 43.17 & 24.02 & 22.95 & 10.25 \\ 
    %     SAM zero-shot & - &  48.76 & 49.03 & 49.03 & 44.66 \\ %\hline
    %     Full finetuning & \text{\sffamily X}  & 79.08 & 52.69 & 53.76 & 47.83 \\ 
    %     Full finetuning &  \checkmark  & 82.68 & 56.96 & 57.17 & 56.37 \\%  \hline
    %     LoRA & \checkmark & 84.52 & 57.56 & 56.44 & 57.14 \\ 
    %     MLP Adapter &  \checkmark & 85.29 & 57.73 & 57.25 & 55.81 \\ 
    %     Prompt Tuning &  \checkmark & 79.31 & 53.27 & 53.49 & 48.87 \\   \hline
    % \end{tabular}
    }
    \caption{\textbf{Performance Evaluation on Pseudo New Modalities.} We combine natural images with a novel image modality as a pseudo new modality: note that we do not use the information that which three channels are for natural images and which channels are for new modalities. SimCMF achieves better performance compared with ``training from scratch" and ``zero-shot". Zero-shot: we use the natural image in the pseudo new modality as the input to the pretrained SAM.}
    \label{tab:comparison_c}
\end{table}

\vspace{\baselineskip}
\noindent\textbf{AIMS dataset construction.}
\label{sec:AIM-Benchmark}
Since there is no existing benchmark that covers different types of modalities for the promotable segmentation task of SAM, we construct a new benchmark named Any Image Modality Segmentation (AIMS) benchmark. Specifically, we choose five representative sensors in different fields and their corresponding images as follows:
\begin{itemize}
    \item \textit{Polarization Images} capture the polarization state of the light. The polarization image is a nine-channel image. The polarization state is closely related to the shape and materials of objects and can be used for challenging tasks for conventional intensity cameras, such as camouflaged object detection, transparent object segment, reflection removal, $etc$.  We adopt RGBP-Glass~\cite{glass_dataset} and ZJU-RGBP~\cite{zjurgbp} in our benchmark. RGBP-Glass contains 3207 and 1304 images for training and evaluation, respectively. ZJU-RGBP includes 344 training images and 50 validation images.   
    % Besides, the polarization camera can also capture the intensity of the scene.
    \item \textit{Depth Images} capture scene geometry, which is commonly used in diverse applications, including robotics, autonomous driving, and computational photography. The depth image captured from the camera is a one-channel image. In our benchmark, we adopt the public NYUv2 dataset~\cite{SilbermanECCV12}, which contains 1449 RGBD samples covering 40 categories. 
    
    \item \textit{HHA Images} are processed features obtained from depth images, which we analyze as a new modality~\cite{gupta2014learning}. The HHA encoding is a method for representing depth images in a way that captures additional geometric information beyond just depth. HHA uses three channels at each pixel to encode the horizontal disparity, the height above ground, and the angle with gravity.

    \item \textit{Thermal Images} capture thermal radiation coming from scenes or environments despite the weather and illumination conditions,
    % perceive the thermal information of scenes or environments, 
    which are commonly in various areas. The thermal images are usually one-channel. In our benchmark, we adopt the public Thermal-based glass segmentation dataset~\cite{huo2023glass}, which contains 5551 images with segmentation labels.
    
    \item \textit{NIR Images} can capture the light in near-infrared frequency, which are commonly used in low-light vision. The NIR (Near-Infrared) images are usually one-channel. We adopt the IVRG-NIR dataset~\cite{brown2011multi} in our benchmark, which consists of 477 NIR images and their ground truth.
    % for comparison experiments, we can refer to the results in ``RANUS: RGB and NIR Urban Scene Dataset for Deep Scene Parsing -- TABLE II".
    % \item \textit{Ultrasound images} are adopted in medical imaging and defect detection.
    % {(1) What is it? (2) Applications (3) Details (which dataset)}.  Near-infrared\textcolor{red}{(1) What is it? (2) Applications (3) Details (which dataset)}.  Near-infrared\textcolor{red}{(1) What is it? (2) Applications (3) Details (which dataset)}
\end{itemize}

% \begin{figure}[ht!]
%   \centering
%   \includegraphics[width=0.9\linewidth]{Figure/baseline_comp_v2.pdf}
%     \caption{\textbf{Performance Evaluation on Different Modalities.} {\textbf{a}.} The proposed method SimCMF improves the segmentation performance significantly on all evaluated modalities compared with training the models from scratch. Specifically, SimCMF improves the mIoU from 22.15\% to 53.88\% for all evaluated modalities on average.  Besides, the peak performance between finetuning and parameter-efficient finetuning is similar. 
%     \textbf{b}. Results on Pseudo New Modalities. We combine natural images with a novel image modality as a pseudo new modality: note that we do not use the information that which three channels are for natural images and which channels are for new modalities.  For example, our MAT is effective in improving the finetuning performance on all evaluated pseudo new modalities. Besides, the peak performance between finetuning and parameter-efficient finetuning is similar.
%     \textbf{c}. We provide controlled experiments for different finetuning strategies on new modalities. Parameter-efficient finetuning strategies can achieve comparable performance compared with full finetuning by using much less trainable parameters.
%   }
%   \label{fig:baseline_cmp_num}
% \end{figure}

We select these modalities as they capture significantly different properties of scenes compared with conventional intensity cameras, and they are quite different from each other. Besides, there are publicly available segmentation datasets for these modalities, and the effectiveness of the novel modality has been proven in previous works. Compared with the training data of RGB-based SAM, which contains 11 million images and more than 1 billion masks, most datasets have a limited number of training images and masks. The segmentation labels of SAM are instance-level segmentation. However, for some segmentation datasets, only semantic labels are provided, which is different from the requirement of the SAM training setting. Hence, post-processing is required to convert the ground truth format to the SAM training setting. Details are presented in the Supplementary Information.
\vspace{\baselineskip}

\noindent\textbf{Evaluation metric.} We evaluate SimCMF for segmentation transfer across modalities on our constructed dataset. Following the protocol of the interactive setting adopted in SAM~\cite{kirillov2023segment}, the center point of an instance is used as the default click prompt fed into the network. We adopt ViT-Base as the image encoder backbone of the pretrained SAM for all experiments. As the best learning rate can be different for each model, we sweep the learning rates and report the best performance for each model for a fair comparison. For all evaluated modalities, we only require the number of channels $C$ and then build our SimCMF for end-to-end training.

\begin{figure}[t!]
\centering
\hspace*{-16pt}\includegraphics[width=1.14\linewidth]{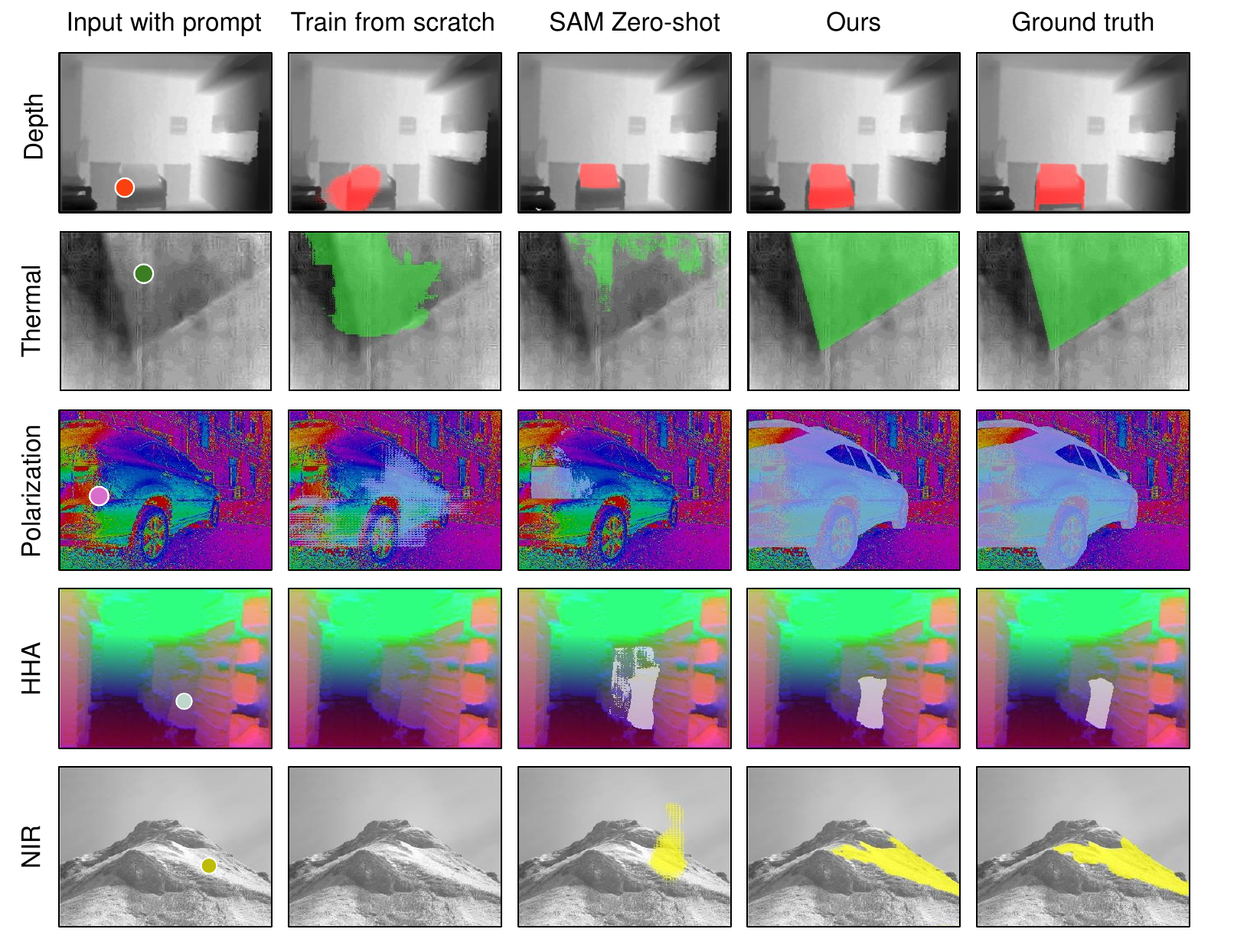}
\vspace{-1em}
\caption{\textbf{Qualitative Results.} We transfer the segment anything ability of SAM to different modalities, including segmentation from depth, thermal, polarization, HHA, and NIR images. The proposed method significantly improves segmentation quality compared to SAM zero-shot and training from scratch. }
\label{fig:qualitative_result}
\end{figure}

\subsection{Performance evaluation}

\paragraph{Comparison to training from scratch.} Before exploring how to perform cross-modal fine-tuning, we first implement baseline approaches as references, $i.e.$, \textit{training from scratch}. The most naive baseline is to inherit the SAM architecture without pretrained weights and train the network only with the new modality data. While we understand it is quite challenging to train a Transformer model effectively with a small amount of data, we adopt this method as a baseline to keep experimental factors the same for reference. This baseline approach only achieves a low \exptrainingsctrach \% average mIoU on our benchmark.

Training with SimCMF can achieve significantly better performance compared with training the models on specific data from scratch on our evaluated dataset. Table~\ref{tab:comparison_a} presents the results of our approach on different modalities. The results of training from scratch for all modalities are poor, which only gets \exptrainingsctrach\% mIoU on different modalities. As a comparison, training the model with SimCMF achieves \expSimCMF\% mIoU, which is significantly better than training from scratch. This phenomenon is according to our expectations as the transformer is data-hungry and requires a large number of data for training. Since the dataset size for these sensors is usually small, they cannot train a good foundation model. This significant improvement demonstrates the potential and importance of cross-modal fine-tuning from vision foundation models to other modalities in different fields. 
We further analyze the visual results to better understand the phenomenon in Figure~\ref{fig:qualitative_result}. The perceptual performance of training from scratch is poor, where the mask is inaccurate. As a comparison, training the model with SimCMF can obtain accurate and sharp segmentation results. We observe similar phenomena in all evaluated modalities, which demonstrates the effectiveness of our proposed SimCMF.
\vspace{\baselineskip}

\noindent\textbf{Experiments on pseudo new modalities.} We further study a data representation that combines natural images and a paired new modality. This representation has been studied in previous methods as a multi-modality representation~\cite{zhu2023visual}.
However, in our setting, while we use both natural images and new modality, we \textit{assume} that we do not know the modality sequence: we do not adopt any RGB prior knowledge to process the RGB images individually so that it could be a special case of new modality named pseudo-new modality. Note that while this prior knowledge is easy to obtain, we just use these pseudo-new modalities to validate the effectiveness of our approach. Specifically, we shuffle the channels to avoid using domain knowledge. Since we have access to RGB images in this experiment, we provide an additional reference method of directly inputting the RGB image to SAM, which we denote as SAM zero-shot. As we can see in Table~\ref{tab:comparison_c}, SAM zero-shot achieves reasonable performance but is far from our approach. As a comparison, our SimCMF framework with different finetuning strategies achieves much better performance compared with baselines. 
% More qualitative results are presented in the supplementary materials due to limited space.
\begin{figure}[t!]
\raggedright
\includegraphics[width=0.95\linewidth]{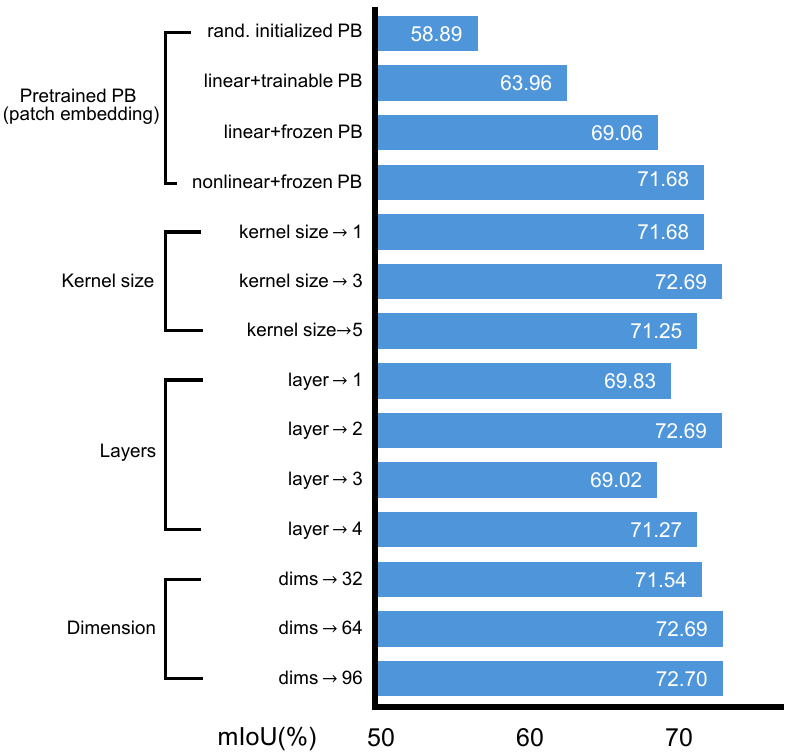}
\vspace{-0.5em}
\caption{\textbf{Exploring Cross-modal Alignment Module.} Randomly initializing a patch embedding for each modality leads to the worst result. A simple linear layer with the pretrained embedding layer can improve the performance already. Interestingly, the results would be better if we frozen the embedding layer. Introducing the nonlinearity is beneficial for the transfer performance. All models are trained with a parameter-efficient fine-tuning strategy. The experiments here are conducted on polarization datasets, and we also validate the effectiveness of these designs on other modalities. }
\vspace{-1em}
\label{fig:MAT_design}
\end{figure}

\subsection{Analyzing the cross-modal alignment module}
\label{subsec:MATNet}
Dimension misalignment is an inevitable challenge in cross-modal fine-tuning. While there are some commonly adopted naive strategies, such as using linear layers to change the dimensions, handling dimension misalignment with satisfying performance is still an open problem to solve up to date. In this section, we provide extensive experiments to show the design process of our cross-modal alignment module. Due to limited computing resources, all experiments are conducted using polarization images~\cite{glass_dataset} that consist of unpolarized intensity images, angle of linear polarization images, and degree of linear polarization images as the target new modality. Polarization images have nine channels. We load the pretrained weights for the foundation model and train the whole model jointly. We adopt LoRA~\cite{hu2021lora} to train these models.

\vspace{\baselineskip}

% Table~\ref{table:MAT_design} shows different existing methods for solving the dimension misalignments in different tasks. There are some direct dimension alignment strategies adopted in priors works. We apply these methods in our SimCMF framework, but they cannot achieve satisfying performance. 

\noindent\textbf{Pretrained vision patch embedding.} We start by replacing the pretrained embedding layer with a randomly initialized embedding layer. This strategy is quite direct and naive, which has been commonly used in prior works~\cite{sun2019rtfnet,singh2023depth,lu2022frozen}. Compared to training from scratch, this implementation fully utilizes the pretrained vision model weights. As a result, mIoU is improved from \mattrainingsctrach \% to \matrandomly \% compared to training from scratch, validating the potential of modality-agnostic transfer learning. 

We then try another common strategy that uses a linear layer and adopts the pretrained patch embedding. This operation improves the performance from 58.89\% to 63.96\%. Inspired by MoCo-v3~\cite{moco-v3}, which observes that patch embedding is critical to the training stability, we try to freeze the pretrained embedding layer in the whole training process. As a result, we observe that freezing pretrained patch embedding does improve the performance from 63.96\% to 69.06\%. 

\textit{We thus use frozen pretrained embedding layer.}

\vspace{\baselineskip}
\noindent\textbf{Nonlinearity.} While using a linear layer is simple and effective, we suspect the transformation of the linear layer is too simple to align two different image modalities, preventing it from achieving better performance. Introducing nonlinearity and replacing the linear layer with MLP is adopted in works like vision-language-model~\cite{llava} and contrastive learning~\cite{chen2020simple,moco-v3}. We add a ReLU layer and another linear layer to train the model again. We notice the performance can be improved from 69.06\% to 71.68\%. 

\textit{We thus adopt the nonlinearity.}

\vspace{\baselineskip}

\noindent\textbf{Kernel size.} Different from other modalities, images have abundant local features. Hence, it is worthwhile to analyze that is $1\times1$ convolution (i.e., linear layer) the best choice. We thus set the number of convolution layers to 1,3,5 to explore this question. As a result, we notice that setting the kernel size to 3 achieves the best performance, which obtains 72.69\% mIoU on our evaluated dataset.

\textit{We thus set the kernel size to 3.}

\vspace{\baselineskip}

\noindent\textbf{Number of layer and dimension.} Increasing the number of learnable parameters is usually an effective strategy with an appropriate training strategy. However, adding the initialized parameters before the foundation model backbone will affect the output significantly. Hence, it is still an open problem that: \textit{can we achieve better performance by adding more parameters before loading the pretrained embedding layer and foundation model bacobone?} To answer this question, we first sweep the number of layers from 1 to 4. We observe that using more than two layers cannot obtain better performance. Then, we sweep the dimensions of convolution layers from 32 to 96. Nevertheless, increasing the dimension from 64 to 96 can only slightly improve the performance. 

\textit{
We thus use two layers and set the dims to 64. 
}

\vspace{\baselineskip}

\noindent\textbf{Discussion on other alternatives.} With extensive experiments, we propose a very simple yet effective design for aligning the dimension misalignment. There are some other designs for the cross-modal alignment adapter and we discuss them here to provide insights to the community. First, we try to replace the convolution layers with transformer blocks before the pretrained embedding layer. However, it can hardly obtain more than 30\% mIoU by changing the hyper-parameters extensively on our evaluated setting, which is significantly worse than our designed cross-modal adapter. Besides, we also compare another strategy used in MedicalSAM~\cite{wu2023medical}, which transposes the feature dimension to the batch dimension and processes them separately. Specifically, it gets 70.90\% on our evaluated dataset, which is close to our results, but it suffers from a practical resource problem: for single-channel images, the FLOPs of this method are similar to ours; however, it uses around 9$\times$ FLOPs compared with our cross-modal adapter when using polarization images. At last, note that while some representative works like ORCA~\cite{shen2023cross} and BLIP-2~\cite{li2023blip} propose methods for aligning dimensions, they require source-target paired data (e.g., image-text pair), which are not available under our setting. Hence, their modules are not applicable. 
\begin{table}[t!]
    \centering
    \resizebox{1.0\linewidth}{!}{
    \begin{tabular}{lcccccc}
        \hline
        Method &  Thermal & Depth & HHA & NIR & Polarization \\ \hline
        % Full finetuning  & 53.76 & 41.12 & 40.78 & 40.99 & 62.31 \\ %\hline
        Full finetuning & 57.17 & 43.98 & 44.02 & 49.79 & 69.19 \\ 
        LoRA  & 56.44 & 45.09 & 43.40 & 52.66 & 72.69 \\ 
        MLP Adapter  & 57.25 & 44.36 & 42.88 & 50.36 & 71.83 \\ 
        Prompt Tuning &  52.78 & 40.92 & 38.79 & 46.09 & 64.08 \\ \hline
    \end{tabular}
    % \begin{tabular}{lccccccc}
    %     \hline
    %     Method & MAT & Thermal & Depth & HHA & NIR & Polarization \\ \hline
    %     Full finetuning &  \text{\sffamily X}  & 53.76 & 41.12 & 40.78 & 40.99 & 62.31 \\ %\hline
    %     Full finetuning & \checkmark & 57.17 & 43.98 & 44.02 & 49.79 & 69.19 \\ 
    %     LoRA & \checkmark & 56.44 & 45.09 & 43.40 & 52.66 & 72.69 \\ 
    %     MLP Adapter & \checkmark & 57.25 & 44.36 & 42.88 & 50.36 & 71.83 \\ 
    %     Prompt Tuning & \checkmark &  52.78 & 40.92 & 38.79 & 46.09 & 64.08 \\ \hline
    % \end{tabular}
    }
    \caption{\textbf{Experiments on Different Finetuning Strategies.} Parameter-efficient finetuning strategies can achieve comparable performance compared with full finetuning by using much less trainable parameters (4\% $v.s.$ 100\%).}
    \label{tab:comparison_b}
\end{table}
\subsection{Empirical analysis of fine-tuning strategies}
\label{sec:finetuning}

\begin{figure*}[t!]
\vspace{-12pt}
  \centering
  \includegraphics[width=1\linewidth]{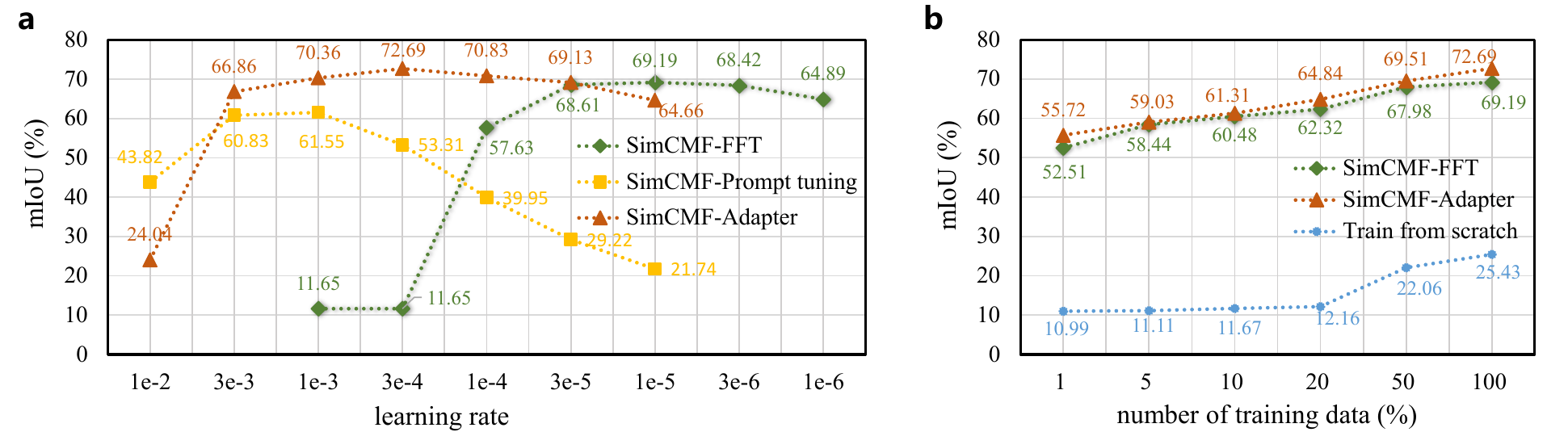}
  \vspace{-1.5em}
    \caption{\textbf{The Effect of Learning Rate and Training Data Size.} The models are evaluated on the polarization modality. \textbf{a}. the full fine-tuning and parameter efficient tuning achieve peak performance in different learning rates. \textbf{b}. increasing the scale of training data brings consistent performance improvement across different training strategies.}
  \vspace{-1em}
  \label{fig:lr_data_effect}
    \end{figure*}

After handling the dimension misalignment with our designed cross-modal alignment module, SimCMF can adopt existing fine-tuning strategies easily like unimodal fine-tuning. However, fine-tuning strategies are not validated systematically on many sensors in different areas, such as polarization~\cite{lei2020polarized}. We mainly explore two commonly used fine-tuning styles here: (1) Full Fine-tuning (FFT): full fine-tuning is commonly used as it usually achieves satisfying performance easily. (2) Parameter-efficient Fine-tuning (PEFT): as the parameters of foundation models are usually very large, full fine-tuning a model might be extremely resource-hungry. PEFT strategies usually fix the original parameters and introduce a small amount of learnable new parameters. In our experiments, we select representative methods, including LoRA, MLP Adapter, and prompt tuning.

 % Following this setting, we make both the MAT layer and pretrained backbone learnable.

Table~\ref{tab:comparison_b} presents the results of different fine-tuning strategies. We sweep the learning rate for each fine-tuning method and choose the best result for comparison. All fine-tuning strategies can improve the segmentation performance compared with training from scratch. LoRA and Adapter can achieve similar performance with full finetuning while they only use much fewer trainable parameters. The results of prompt tuning fall below that of the other two PEFT methods despite having a close number of trainable parameters, and we believe this is attributed to the initial noise brought by the prompts embedding. It fails to find an initialization that ensures prompt embeddings do not disturb the model output at the first forward pass. In comparison, the effects of both LoRA and MLP adapter on the model can be initialized as zero.

In Figure~\ref{fig:lr_data_effect}(a), we further study the effect of learning rate for different fine-tuning strategies. Prior works~\cite{hu2021lora,jia2022visual} noticed that different tuning strategy holds different best learning rates; we observe consistent results here. 
Full fine-tuning achieves a peak performance 69.19\% mIoU at lr=1e-5, while parameter-efficient finetuning achieves the best 72.69\% mIoU at lr=3e-4. 
We suspect the reason is the number of trainable parameters. Full fine-tuning makes all parameters learnable; a small learning rate prevents the model from deviating far away from the pretrained weights. 
While LoRA or MLP Adapter with only 4\% trainable parameters demands a larger learning rate for efficient learning. 

In Figure~\ref{fig:lr_data_effect}(b), we study the relationship between the number of finetuned images and the pretrained model. We split the training set randomly according to different ratios. shows the results. We notice that using RGB-based pretrained SAM can significantly improve the performance on different image modalities, especially when the training images of specific modalities are limited.

\section{Conclusion}
Foundation models (Large Models) have revolutionized artificial intelligence areas, such as ChatGPT~\cite{ouyang2022training} in natural language processing and SAM (Segment Anything Model) in computer vision. Driven by the availability of large-scale image data, several foundation models have recently been proposed~\cite{clip,controlnet,kirillov2023segment,lisa} for vision tasks, including image recognition, segmentation, conditional generation, etc. As a result, numerous downstream tasks can achieve impressive performance. 
Nevertheless, except for conventional cameras, the available data of many image sensors is not large enough, preventing applications in different areas from benefitting from the significant progress of foundation models. Transferring the ability of vision foundation models to new data-limited image modalities is promising, but this line of work has not been fully explored or studied. 

% Our contribution
In this work, we confirm the potential of modality-agnostic cross-modal fine-tuning from vision foundation models to other image modalities beyond natural images. To this end, we introduce a training paradigm, SimCMF, to study this problem. We conduct extensive exploratory experiments to propose a practical cross-modal alignment module for receiving different types of new modalities. We explore different finetuning strategies and report our observations. Based on these experiments, we validate the transfer performance of our proposed SimCMF through a vision foundation model SAM on a variety of sensors. The significant margins achieved by SimCMF suggest that the generic cross-modal transfer learning is still underexplored. We envision SimCMF to be useful for other vision foundation models and other unevaluated modalities that are not studied in this work.

\section{Acknowledgment}
The authors would like to thank Ziwei Liu for his help.

\section{Technical Report}
The previous version of SimCMF is available at \url{https://arxiv.org/pdf/2409.08083}.

{
    \small
    \bibliographystyle{ieeenat_fullname}
    \bibliography{main}
}


\maketitlesupplementary

\section*{Outline}
\label{sec:benchmark}

This supplementary document provides further description and additional results to support the findings from
the main manuscript. The document is organized as follows.

\begin{itemize}
    \item \textbf{Section~\ref{sec:add_training}}: This section provides additional training details of SimCMF.
    \item \textbf{Section~\ref{sec:add_exp}}: This section provides additional controlled experiments.
    \item \textbf{Section~\ref{sec:add_comp}}: This section provides additional comparison results.
    \item \textbf{Section~\ref{sec:add_bench}}: This section provides additional detailed description on building benchmark.
    \item \textbf{Section~\ref{sec:add_qualitative}}: This section provides additional qualitative visualization.

\end{itemize}

\section{Additional Training Details}
\label{sec:add_training}
We report the effect of different finetuning strategies on trainable parameters in Table~\ref{tab:params}. The foundation model SAM with ViT-B~\cite{dosovitskiy2020image} as backbone contains 93.7M parameters from the image encoder, prompt encoder, and mask decoder. 
Full finetuning makes all parameters trainable. 
For parameter-efficient tuning, we implement four typical methods including LoRA~\cite{hu2021lora}, MLP adapter~\cite{houlsby2019parameter}, prompt tuning~\cite{jia2022visual}, and full finetuning. Following He et al.~\cite{he2021towards}, we balance their trainable parameters to achieve approximately 4\% of full parameters for fair comparison.

The detailed training configuration is presented in Table~\ref{tab:config}. 
We fix the training epoch to 50 and set the batch size as 4 regardless of the number of training samples in different modality datasets.
We sweep the learning rates from 3e-6 to 3e-3 and report the peak performance as the final result. 
The input modality images are resized to (1024, 1024) to meet the requirements of SAM.

\begin{table}
\begin{center}
\renewcommand{\arraystretch}{1.1} % 增加行高
\resizebox{1\linewidth}{!}{
\begin{tabular}{l|c}
    \specialrule{1pt}{0pt}{0pt}
      Finetuning Strategies & Trainable Parameters (M) \\
       & of Foundation Model  \\
       \hline
    LoRA &  4.3  \\ 
    MLP adapter & 3.9 \\
    % Prefix tuning & 4.4 \\
    Prompt tuning & 4.4 \\
    Full finetuning & 93.7 \\
    \specialrule{1pt}{0pt}{0pt}
    
\end{tabular}
}
\end{center}
% \vspace{-1.2em}
\caption{ \textbf{The Number of Trainable Parameters in Foundation Model (SAM) with Different Finetuing Strategies.} Three parameter-efficient finetuning methods hold similar trainable parameters, which are much less than the trainable parameters of full finetuning strategies.}
\label{tab:params}
% \vspace{-0.5em}
\end{table}

\begin{table}
\begin{center}
\renewcommand{\arraystretch}{1.1} % 增加行高
\resizebox{1\linewidth}{!}{
\begin{tabular}{l|c}
    \specialrule{1pt}{0pt}{0pt}
      Config & Value \\
     \hline 
     optimizer  & Adam  \\
     optimizer momentum &  $\beta_1$,$\beta_2$=0.9,0.999 \\
     batch size & 4 \\
     epoch & 50 \\ 
     learning rate & \{3e-6, 1e-5, 3e-5, 1e-4, 3e-4, 1e-3, 3e-3\} \\
     learning rate schedule & step decay \\
     schedule step size & 10 epoch \\
     schedule gamma & 0.5\\
     augmentation & \text{Resize}(1024, 1024) \\
      \specialrule{1pt}{0pt}{0pt}
    
\end{tabular}
}
\end{center}
% \vspace{-1.2em}
\caption{ \textbf{The Training Setting for Our Experiments.} }
\label{tab:config}
% \vspace{-0.5em}
\end{table}

% \begin{table}[t]
% \begin{center}
% \renewcommand{\arraystretch}{1.1} % 增加行高
% \resizebox{0.95\linewidth}{!}{
% \begin{tabular}{l|cc|ccc}
%     \specialrule{1pt}{0pt}{0pt}
%     \multirow{2}{*}{Model} & \multirow{2}{*}{Modality} &  \multirow{2}{*}{Channels}  &   \multicolumn{3}{c}{Trainable Parammeters (M)}  \\
%       & &   & \small LoRA & \small Adapter &  \small FFT   \\ 
%     \specialrule{1pt}{0pt}{0pt}
%     \rowcolor{gray!50} \small SAM & \small RGB  & \small 3 & \small 4.3 & \small 3.9  & \small 93.6  \\ 
%     \hline 
     
%      \multirow{3}{*}{SimCMF} & \small Depth/Thermal/NIR & \small 1 & \small 4.5  & \small 4.1  & \small 93.8 \\
%      & \small Depth/Thermal/NIR+RGB & \small 4 &  \small 5.1 & \small 4.7  & \small 94.4 \\
%      & \small Polarization & \small 9 & \small 6.0 & \small 5.7 & \small 95.4 \\
%     \specialrule{1pt}{0pt}{0pt}
    
% \end{tabular}
% }
% \end{center}
% % \vspace{-1.2em}
% \caption{ \textbf{The parameters comparison of foundation model (SAM) and SimCMF.} SimCMF only introduces an approximate 0.2\%-2\% increase in parameters.}
% \label{table:params}
% % \vspace{-0.5em}
% \end{table}

% SimCMF can handle modality with arbitrary numbers of channel inputs, its parameters adapt accordingly to different inputs. 
% We present the detailed trainable parameters of SimCMF across different modalities and tuning methods in Table~\ref{table:params}.
% Compared to foundation model SAM, our SimCMF as a lightweight module introduces approximately 2\% extra trainable parameters.

\section{Additional Controlled Experiments}
\label{sec:add_exp}
We provide the study of the hyper-parameter setting of SimCMF by applying it to the Polarization modality. As shown in Figure~\ref{fig:effect}, SimCMF stack the \textit{n} convolutional layers with \textit{k} kernel size and dimension \textit{d}. SimCMF achieves best 72.7\% mIoU by setting {\textit{n}, \textit{k}, \textit{d}\} as  \{2, 3, 64\}. Further increasing the number of stacked layers and dimensional does not bring additional improvements, we suspect it is caused by the factor that introducing more trainable parameters makes training of SimCMF more challenging.
Note that when the kernel size is set to 1 and layers are set to 2, the approach becomes equivalent to employing an MLP layer adopted in contrastive learning~\cite{chen2020simple,moco}. When the kernel size is set to 1 and layers are set to 1, the implementation becomes equivalent to a linear layer. One can observe that setting the kernel size to 3 achieves peak performance with the best tradeoff between the receptive field and trainable parameters.

\begin{figure*}[t!]
\centering
\begin{minipage}{0.3\textwidth}
\centering
\resizebox{1\textwidth}{!}{
\begin{tabular}{l|ccc}
\hline
\textit{k} & 1 & 3 & 5  \\
\hline
mIoU(\%) & 71.7 & 72.7 & 71.3 \\
\hline
\end{tabular}
}
\end{minipage}
\hfill
\begin{minipage}{0.3\textwidth}
\centering
\resizebox{1\textwidth}{!}{
\begin{tabular}{l|ccc}
\hline
\textit{d} & 32 & 64 & 96 \\
\hline
mIoU(\%) & 71.5 & 72.7 & 72.7 \\
\hline
\end{tabular}
}
\end{minipage}
\hfill
\begin{minipage}{0.35\textwidth}
\centering
\resizebox{1\textwidth}{!}{
\begin{tabular}{l|ccccc}
\hline
\textit{n} & 1 & 2 & 3 & 4 & 5 \\
\hline
mIoU(\%) & 69.8 & 72.7 & 71.1 & 71.3 & 71.8 \\
\hline
Params(K) &  0.03 & 5.4 & 42.3 & 79.3 & 116.2 \\ 
\hline
\end{tabular}
}
\end{minipage}%
\vspace{1mm}

\begin{minipage}{0.3\textwidth}
\centering
    The effect of kernel size.
\end{minipage}
\begin{minipage}{0.3\textwidth}
\centering
    The effect of dimension.
\end{minipage}
\begin{minipage}{0.35\textwidth}
\centering
    The effect of layers.
\end{minipage}

\caption{\textbf{The Effect of the Configuration of our cross-modal alignment module, evaluated on Polarization modality.} Based on the above results, we set the $k,d,n$ to 3, 64, and 2, respectively, considering the trade-off of performance of efficiency.}
\label{fig:effect}
\end{figure*}

\begin{figure*}[t!]
  \centering
  \includegraphics[width=1\linewidth]{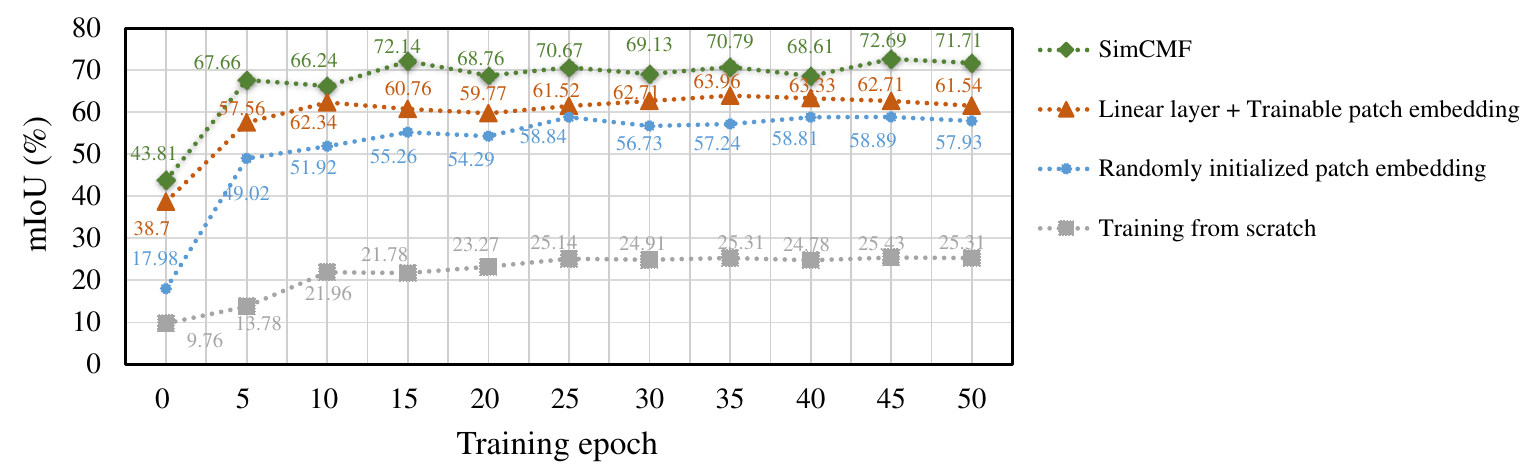}
  \caption{\textbf{The Training Curves for SimCMF and Baselines.} SimCMF achieves the best performance.}
  \label{fig:supp_curve}
\end{figure*}

\begin{table*}[t!]
\small
\centering
\renewcommand{\arraystretch}{1.2}
\resizebox{0.85\linewidth}{!}{
\begin{tabular}{@{}l@{\hspace{3mm}}c@{\hspace{3mm}}l@{\hspace{3mm}}c@{\hspace{3mm}}c@{\hspace{3mm}}c@{\hspace{3mm}}c@{\hspace{3mm}}c@{\hspace{3mm}}c@{\hspace{3mm}}c@{\hspace{3mm}}c@{\hspace{3mm}}c@{}}
\toprule[1pt]

Method &  Params & Finetuning methods & RGB-T & RGB-D & RGB-HHA & RGB-NIR \\
    \midrule 
% Train from scratch & &29.94 & 21.44 & 25.11 & 13.80 \\
% SAM zero-shot & &    48.29 & 48.82 &48.26 & 44.80 \\
 CMX*~\cite{zhang2023cmx} & 403.8M & Full finetuning & 44.91 & 36.41 & 37.33 & 34.75 \\
 ViPT*~\cite{zhu2023visual} & 94.5M& Prompt tuning & 75.93 &48.89 &49.50 & 51.90 \\
\hline
% Med-Adapter & &41.09 &27.49 & & &- &- &- \\
SimCMF & \multirow{3}{*}{94.4M}  & LoRA  & 84.52 & 57.56 & 56.44 & \textbf{57.14} \\
SimCMF & & MLP Adapter &\textbf{85.29}  & \textbf{57.73}& \textbf{57.25} & 55.81 \\
SimCMF & & Full finetuning & 82.68 & 56.96 & 57.17 & 56.37 \\

% Naive PEFT1 &  \\
% Naive PEFT1 + Ours projector& \\
% Naive PEFT2 & \\
% Naive PEFT2 + Ours projector& \\
% Naive PEFT3 & \\
% Naive PEFT3 + Ours projector& \\
% *Align and then refine  &  -&-&-&-&-&-&0\\ 
% Ours (Ours projector + PEFT-X)&\\
\bottomrule[1pt]
\end{tabular}
}
\caption{\textbf{Comparison of SimCMF with Other Methods Tackling Pseudo New Modality (RGBX).} While with fewer parameters, SimCMF achieves better performance across four pseudo new modalities.  Note that ViPT and CMX can tackle RGBX only. * means reproduced implementation in SAM.
% \textcolor{red}{Best lr.} 
}
\label{tab:compare_rgbx}
% \vspace{-2em}
\end{table*}

\begin{figure*}[t]
  \centering
  \includegraphics[width=0.7\linewidth]{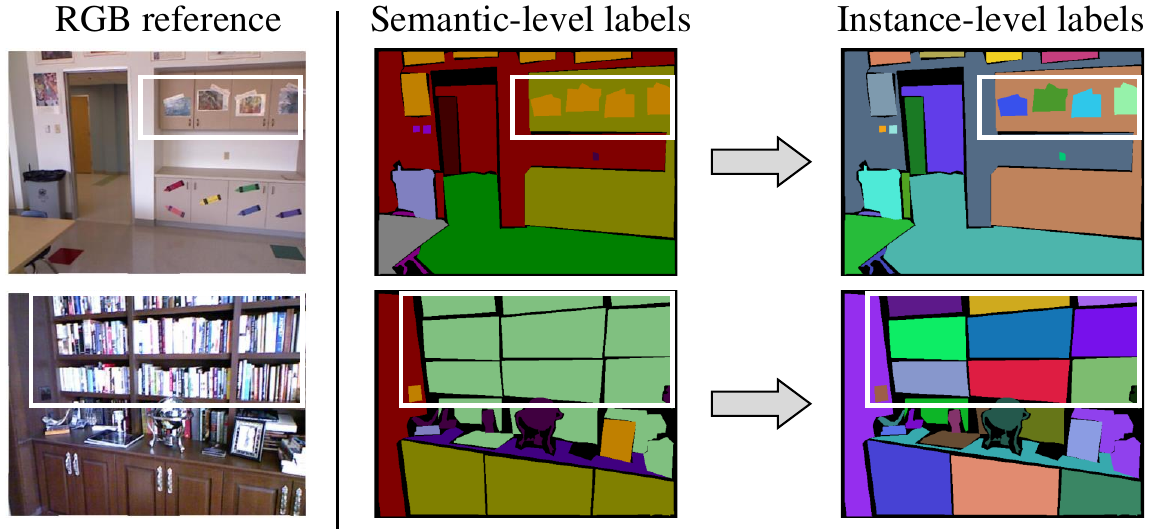}
  % \vspace{-1em}
  \caption{\textbf{The Illustration of Segmentation Generation Pipeline in Our Benchmark.} The semantic-level segmentation ground truth is split into instance-level segmentation ground truth.}
  \label{fig:supp_benchmark}
 % \vspace{-1.5em}
\end{figure*}

\section{Additional Comparisons}
\label{sec:add_comp}
\label{sec:add_baselines}
We report the training curve of SimCMF and baselines on the Polarization dataset in Figure~\ref{fig:supp_curve}. 
One can observe the training from scratch only achieves 25.43\% mIoU, significantly worse than other methods using prestrained weight as initialization. 
To tackle the channel misalignment between RGB modality and new modality input, two straightforward ideas are to build a new randomly initialized patch embedding or prepend a 1$\times$1 convolution layer for dimension projection. While these two methods achieve significant improvement over training from scratch, {their performance is suboptimal.} Our SimCMF achieves a better performance over these two commonly adopted naive baselines. 

Besides, we compare our SimCMF to two SOTA methods with pseudo new modality (RGBX) input. 
\textbf{ViPT}~\cite{zhu2023visual} introduce a modality-complementary prompter (MCP) block to fuse features from RGB and other modalities like thermal and depth.
\textbf{CMX}~\cite{zhang2023cmx} replicate the pretrained RGB encoder to tackle X modality, and place the proposed Feature Rectification Module (FRM) after each block to perform interaction of RGB features and X features. \textit{Note that these two baselines utilize the prior information about which channels are for RGB embedding while our framework does not utilize this information.}
We reimplement the above two methods on SAM following their original finetuning methods and evaluate their performance on our benchmark. 
As shown in Table~\ref{tab:compare_rgbx}, CMX~\cite{zhang2023cmx} does not achieve satisfying performance on finetuning the foundation model SAM. We suspect the unsatisfying performance is caused by the noise introduced from FRM, which appended after each block deviates the features from its original distribution, making the learning difficult. 
While ViPT~\cite{zhu2023visual} can achieve reasonable performance, its performance lags behind SimCMF.

\begin{figure}
  \centering
  \includegraphics[width=0.85\linewidth]{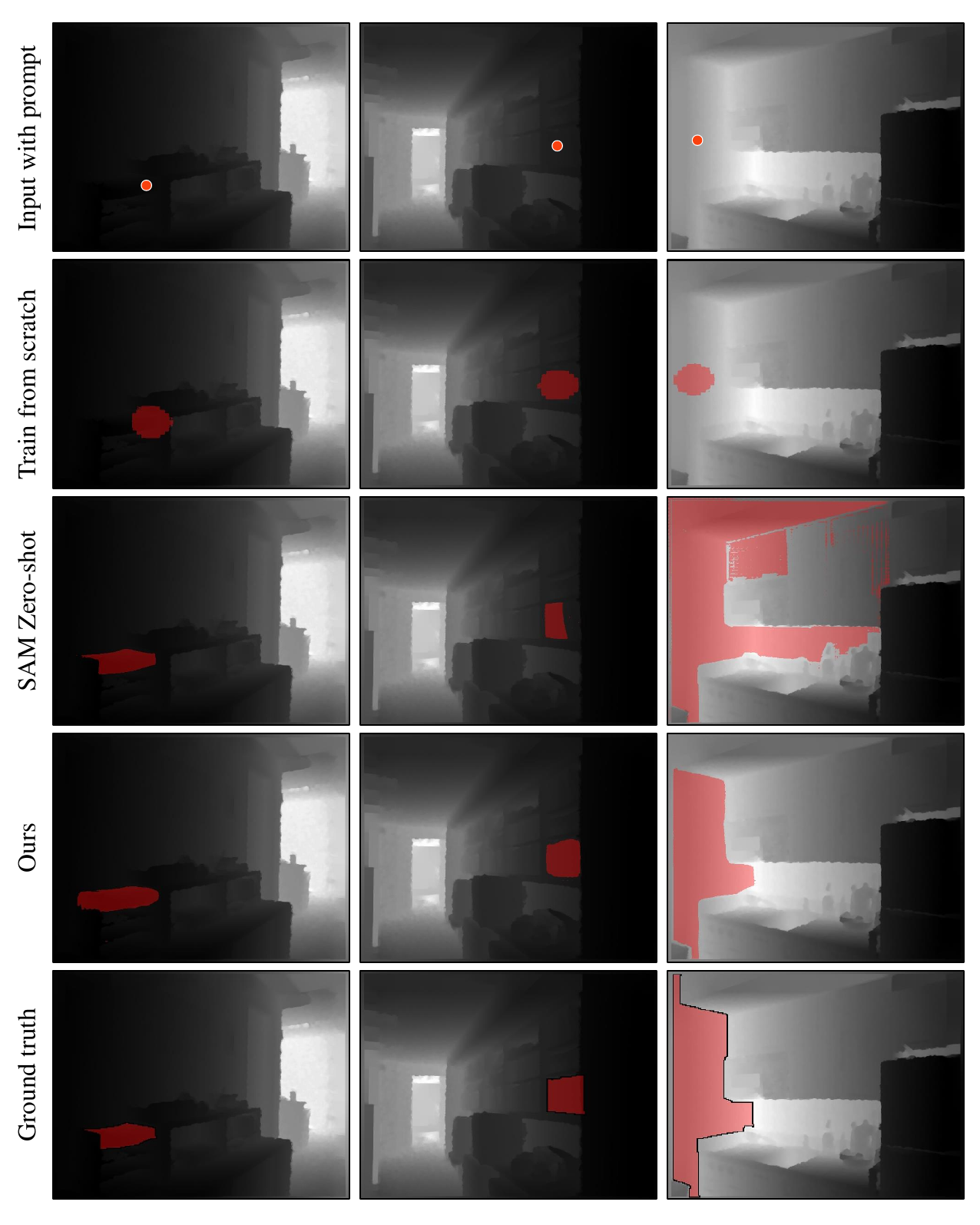}
  \caption{\textbf{Additional Qualitative Results in Depth Modality. } Our approach can perform better than zero-shot and training from scratch. }
  \label{fig:supp_depth}
\end{figure}

\begin{figure}
  \centering
  \includegraphics[width=0.85\linewidth]{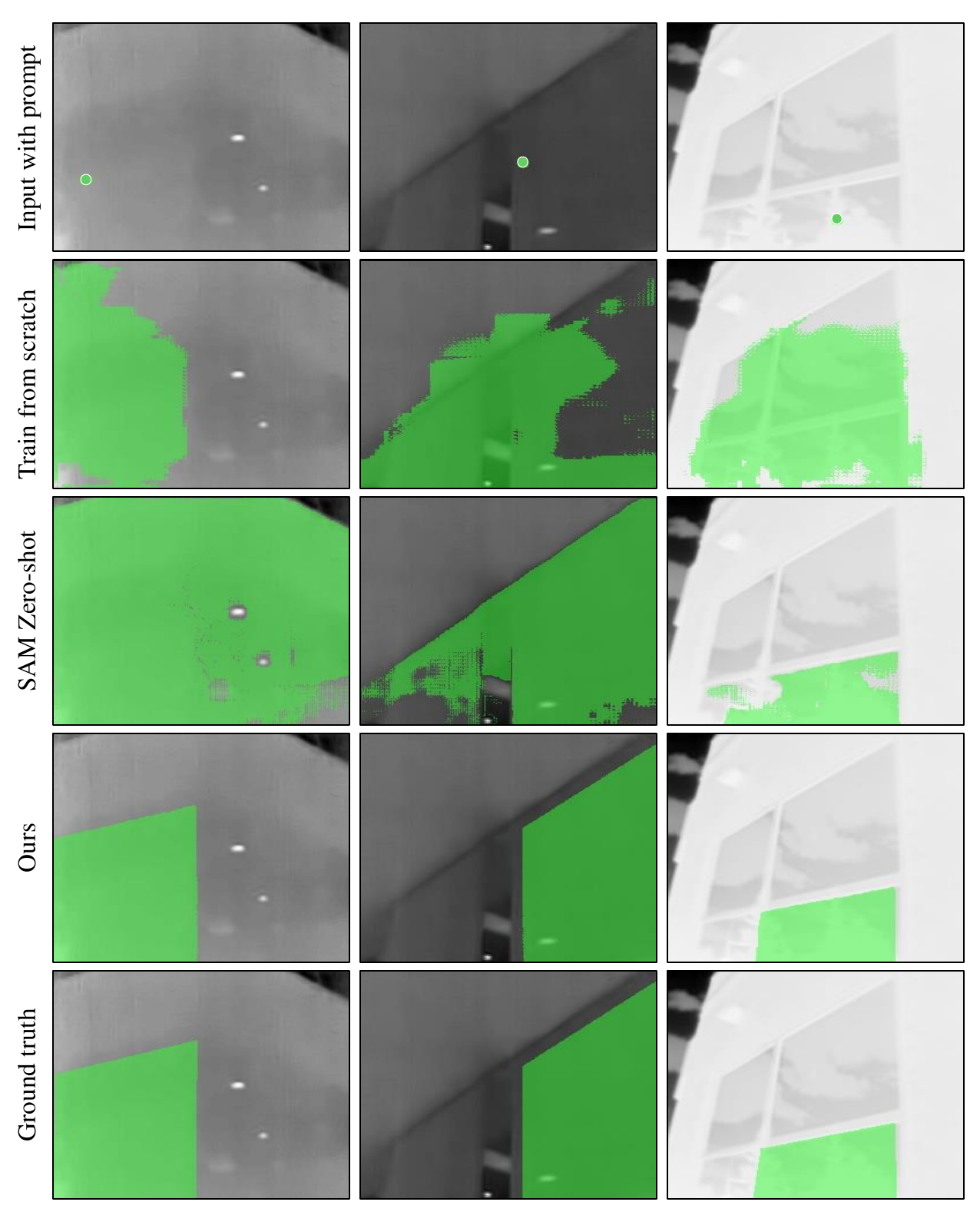}
  \caption{\textbf{Additional Qualitative Results in Thermal Modality. } Our approach can perform better than zero-shot and training from scratch. }
  \label{fig:supp_thermal}
\end{figure}

\begin{figure}
  \centering
  \includegraphics[width=0.85\linewidth]{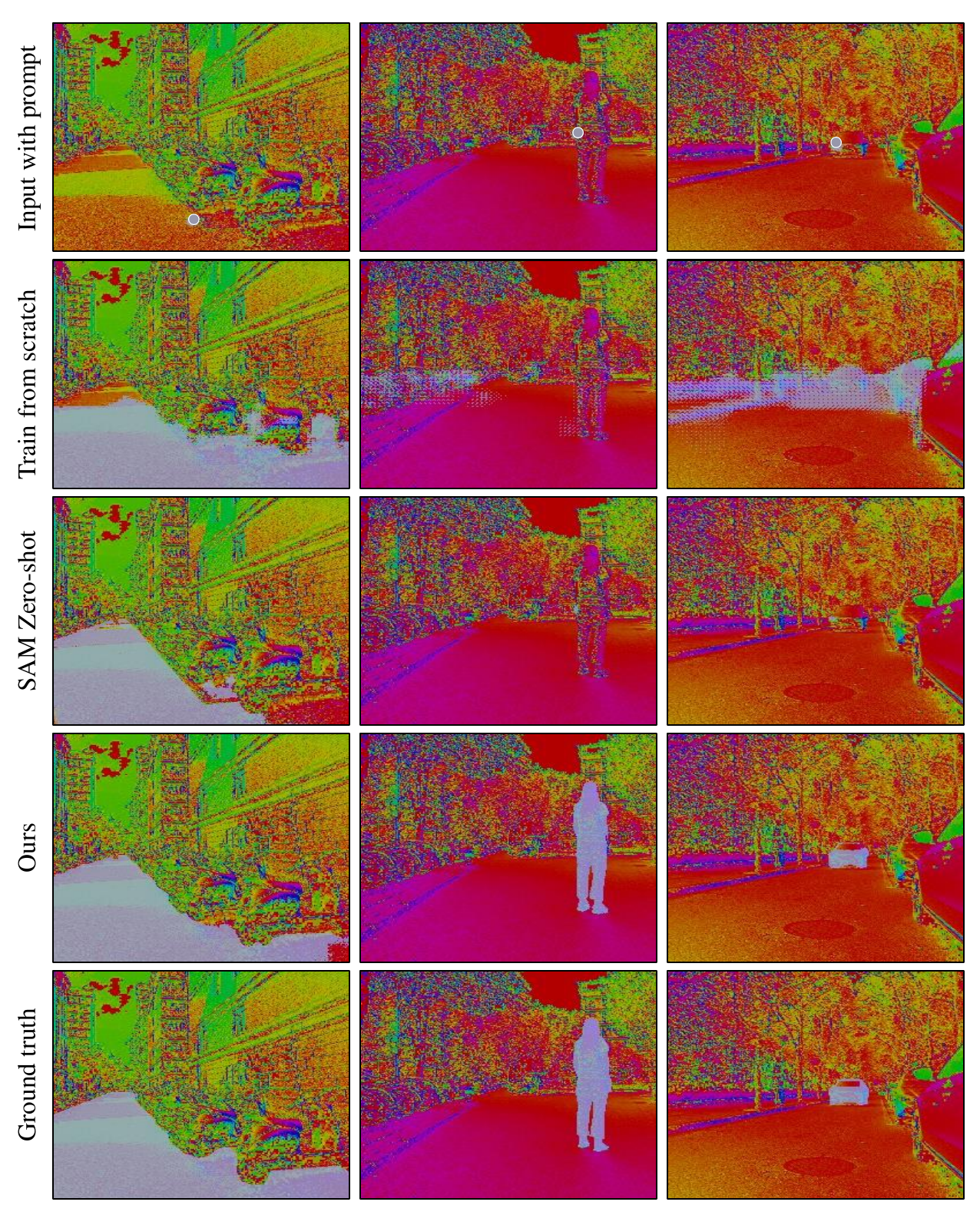}
  \caption{\textbf{Additional Qualitative Results in Polarization Modality. } Our approach can perform better than zero-shot and training from scratch. }
  \label{fig:supp_polarization}
\end{figure}

\begin{figure}
  \centering
  \includegraphics[width=0.85\linewidth]{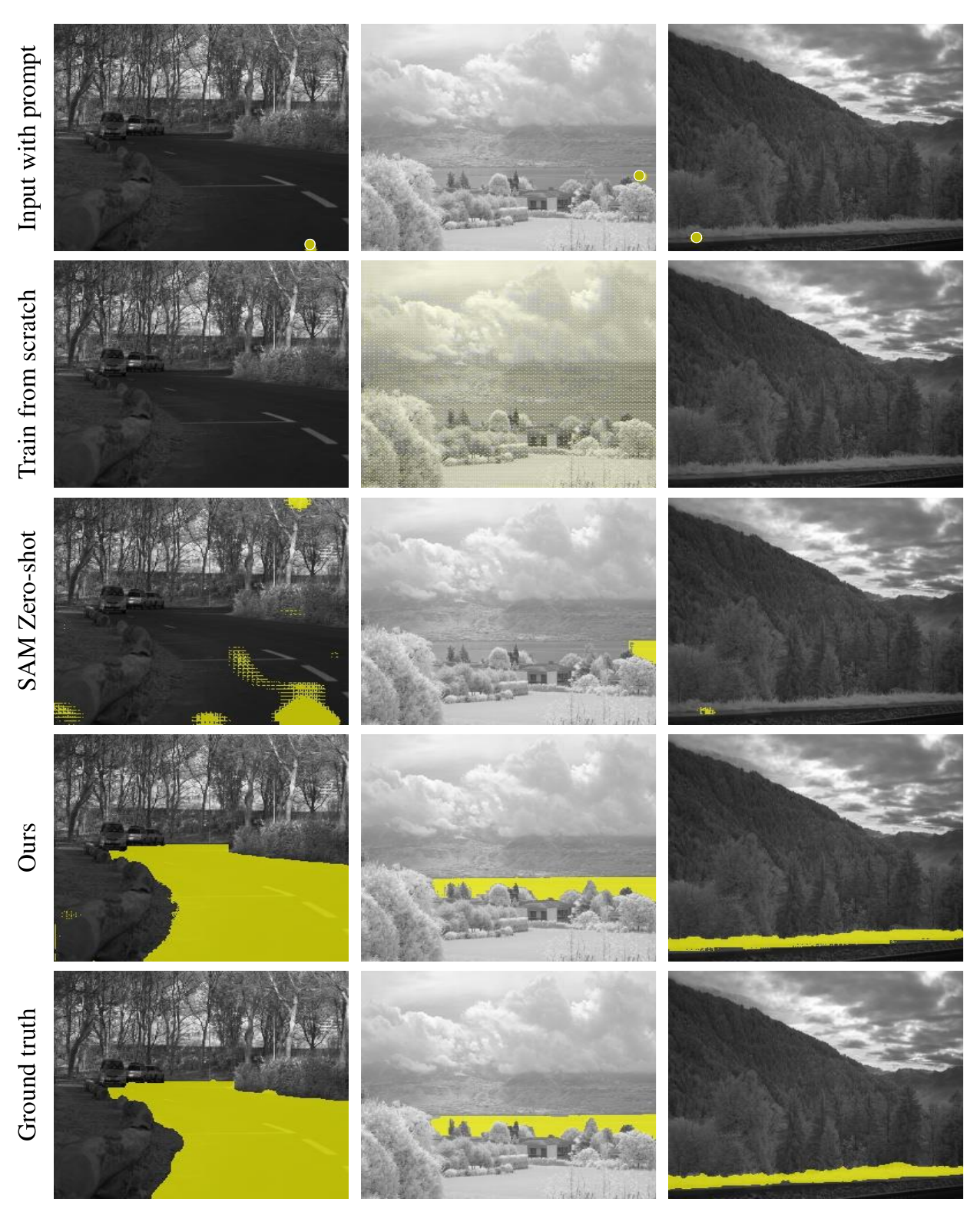}
  \caption{\textbf{Additional Qualitative Results in NIR Modality. } Our approach can perform better than zero-shot and training from scratch. }
  \label{fig:supp_nir}
\end{figure}

\begin{figure}
  \centering
  \includegraphics[width=0.85\linewidth]{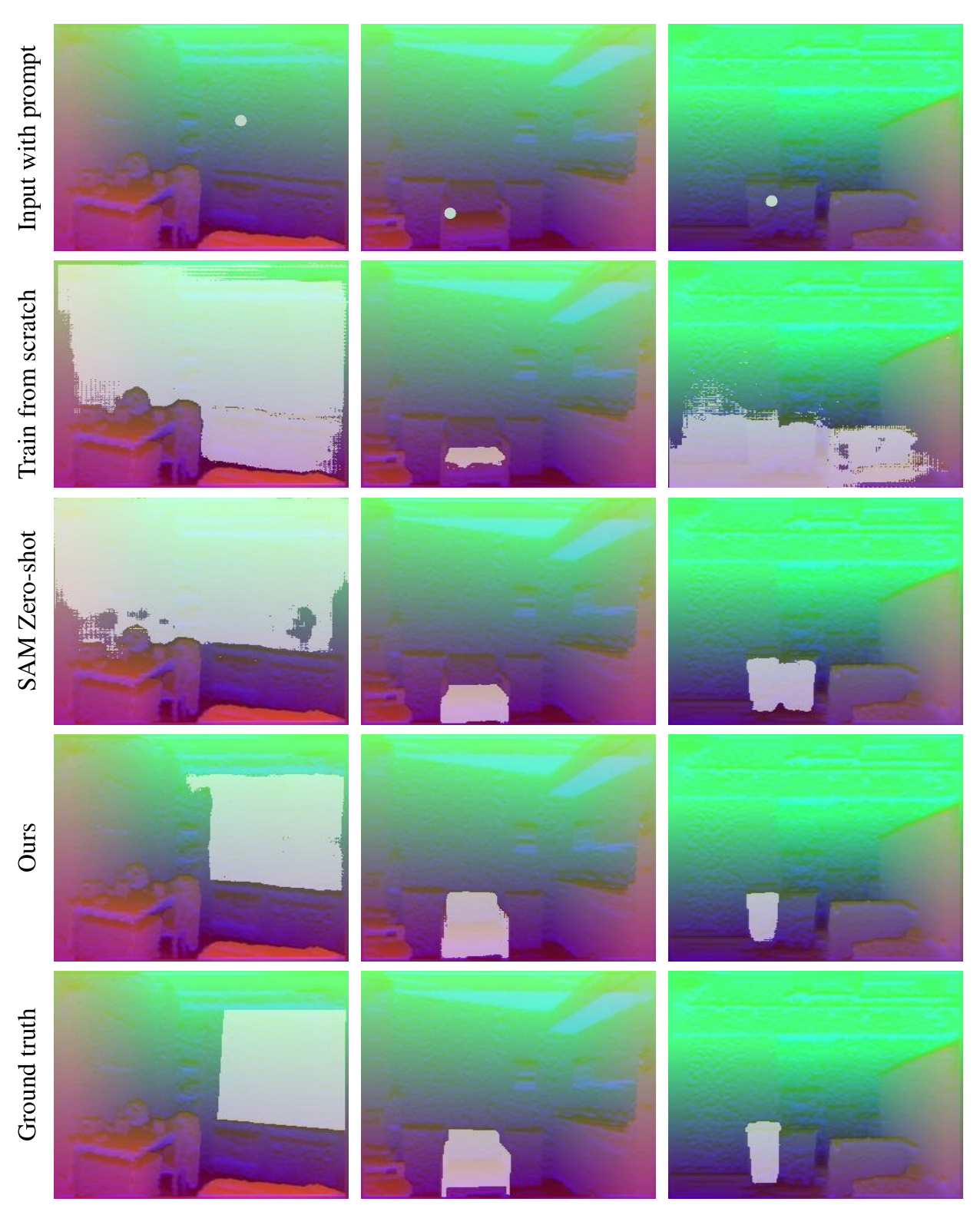}
  \caption{\textbf{Additional Qualitative Results in HHA Modality. } Our approach can perform better than zero-shot and training from scratch. }
  \label{fig:supp_hha}
\end{figure}

\section{Additional Benchmark Details}
\label{sec:add_bench}
To study the problem of cross-modality transfer learning of SAM, we construct a new benchmark by collecting image segmentation datasets from different modalities, as described in the main paper. However, the segmentation labels of SAM are instance-level segmentation, but some segmentation datasets (\textit{e.g.,} ZJU-RGBP~\cite{zjurgbp}, NYUv2\cite{SilbermanECCV12}) only provide semantic labels. Hence, {to align with the output of SAM}, we perform post-processing to convert the semantic labels to instance labels by decomposing non-connected components.

Figure~\ref{fig:supp_benchmark} shows the post-processing effect. Given a semantic map label, we partition it into separate masks if they are not pixel-connected to each other. Each separate mask serves as an instance label and is responsible only for the clicks that lie within it.
The evaluation metric IoU is calculated for each instance. Instead of average IoU over semantic categories, we take the average IoU of all instances as the mIoU results.

\section{Additional Qualitative Results}
\label{sec:add_qualitative}
We provide further qualitative visualizations in Figure~\ref{fig:supp_depth} to Figure~\ref{fig:supp_hha}. For the SAM zero-shot performance, we use the provided RGB reference as the input. We present the results on diverse image modalities for better understanding. As shown in the figure, the performance of training from scratch and zero-shot is generally unsatisfying. {With our proposed SimCMF framework, the segmentation performance can be improved significantly.} 
For example, in the first column of thermal modality in Figure~\ref{fig:supp_thermal}, {we can see that both training from scratch and zero-shot fail to segment the ``window'' completely.} In contrast, our method achieves accurate segmentation, which is quite close to the ground truth mask.

\end{document}